\pdfoutput=1

\documentclass[11pt]{article}

\usepackage[final]{acl}

\usepackage{times}
\usepackage{latexsym}

\usepackage[T1]{fontenc}

\usepackage[utf8]{inputenc}

\usepackage{microtype}

\usepackage{inconsolata}

\usepackage{graphicx}

\usepackage{url}

\usepackage{listings}
\usepackage{xcolor}
\definecolor{pystring}{rgb}{0.164, 0.631, 0.596}    
\definecolor{pycomment}{rgb}{0.498, 0.498, 0.498}   
\definecolor{pykeyword}{rgb}{0.737, 0.353, 0.662}   
\definecolor{pybackground}{rgb}{0.95, 0.95, 0.95}   
\definecolor{pyfunction}{rgb}{0.164, 0.631, 0.596}  
\lstdefinestyle{pythonstyle}{
    language=Python,
    backgroundcolor=\color{pybackground},
    basicstyle=\ttfamily\footnotesize,
    breakatwhitespace=false,
    breaklines=true,
    captionpos=b,
    keepspaces=true,
    numbers=left,
    numbersep=5pt,
    showspaces=false,
    showstringspaces=false,
    showtabs=false,
    tabsize=4,
    frame=single,
    commentstyle=\color{pycomment},
    keywordstyle=\color{pykeyword},
    stringstyle=\color{pystring},
    numberstyle=\tiny\color{pycomment},
    identifierstyle=\color{black},
    emphstyle=\color{pyfunction}
}
\usepackage{xcolor}
\usepackage{pifont}
\usepackage{wasysym}
\newcommand{\cmark}{\textcolor{green}{\ding{51}}}
\newcommand{\xmark}{\textcolor{red}{\ding{55}}}
\newcommand{\pmark}{\textcolor{orange}{\Circle}}

\title{IGC: Integrating a Gated Calculator into an LLM to Solve Arithmetic Tasks Reliably and Efficiently}

\author{Florian Dietz \and Dietrich Klakow \\
  Spoken Language Systems (LSV), Saarland University, Saarbrücken, Germany \\
 \small{
   \textbf{Correspondence:} \href{mailto:fdietz@lsv.uni-saarland.de}{fdietz@lsv.uni-saarland.de}
 }
}

\begin{document}
\maketitle
\footnotetext{This paper is currently under review at ACL Rolling Review.}
\begin{abstract}

Solving arithmetic tasks is a simple and fundamental skill, yet modern Large Language Models (LLMs) have great difficulty with them.
We introduce the Integrated Gated Calculator (IGC), a module that enables LLMs to perform arithmetic by emulating a calculator on the GPU.
We finetune a Llama model with our module and test it on the BigBench Arithmetic benchmark, where it beats the State of the Art, outperforming all models on the benchmark, including models almost two orders of magnitude larger.
Our approach takes only a single iteration to run and requires no external tools. It performs arithmetic operations entirely inside the LLM without the need to produce intermediate tokens. It is computationally efficient, interpretable, and avoids side-effects on tasks that do not require arithmetic operations.
It reliably achieves 98\% to 99\% accuracy across multiple training runs and for all subtasks, including the substantially harder subtask of multiplication, which was previously unsolved.

\end{abstract}

\section{Introduction}

\textbf{Motivation}.
Large Language Models (LLM) have shown impressive abilities in many different fields in recent years \citep{thoppilan2022lamda,chowdhery2023palm,brown2020language}.
This makes it all the more intriguing that even advanced LLMs still perform very poorly on basic arithmetic tasks:
GPT-3 has trouble adding numbers with more than three digits \citep{brown2020language} and GPT-4 \citep{achiam2023gpt} still fails to solve multiplication tasks \citep{dziri2024faith}.
The reasons for this surprisingly poor performance have been studied extensively \citep{yuan2023well,brown2020language,dziri2024faith}.
Even so the BigBench Arithmetic benchmark \citep{srivastava2023beyond}, which tests the four basic arithmetic operations on merely 5-digit long numbers, remains unsolved.

\begin{figure}
    \centering
    \includegraphics[width=0.7\linewidth]{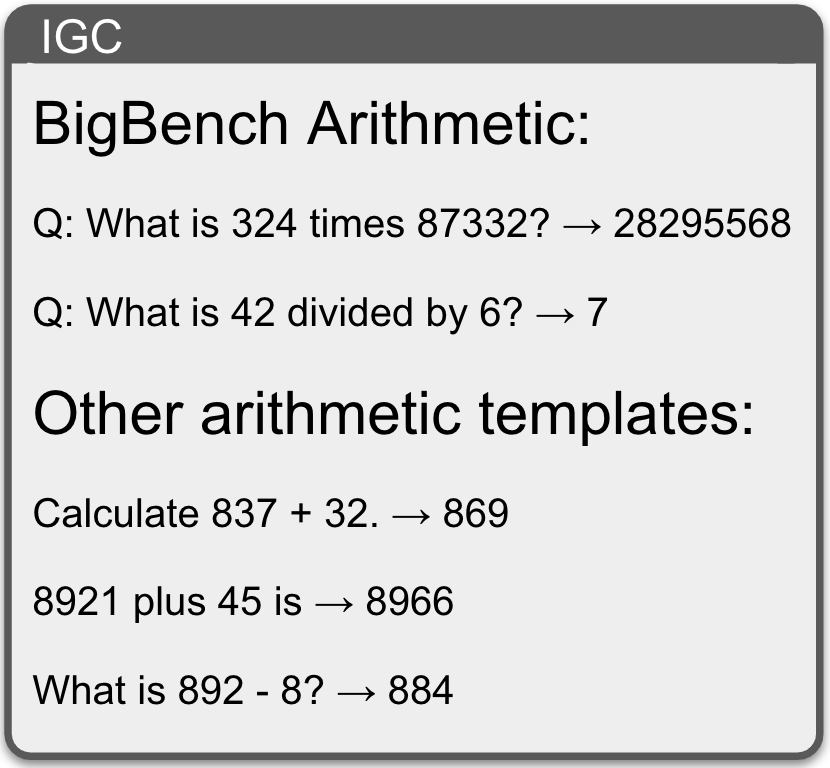}
    \caption{Examples of arithmetic tasks.
    }
    \label{fig:training_templates}
\end{figure}

\textbf{The Impact of Number Representations}.
The arithmetic abilities of LLMs strongly depend on the way numbers are represented \citep{thawani2021representing}.
\citet{mcleish2024transformers} improve arithmetic performance by adding positional encodings to numbers.
Similarly, \citet{liu2023goatfinetunedllamaoutperforms} show that good performance can be achieved on addition and subtraction tasks using finetuning only, and they attribute this to the especially well-suited tokenization method of their model.
However, we are not aware of any finetuning method that solves the more difficult subtask of multiplication effectively.

\textbf{Chain of Thought}.
Chain of Thought (COT) is a prompting method that works by breaking the task down into smaller subtasks and solving them step-by-step \citep{wei2023chainofthoughtpromptingelicitsreasoning}.
This approach makes many difficult tasks solvable, but it also increases the model's runtime as it requires many intermediate outputs to produce the final result.

\textbf{Tool Use}.
\citet{schick2023toolformerlanguagemodelsteach} introduced the Toolformer, which teaches LLMs to call external tools.
This method is very powerful, but it increases the inference time due to costly transfers of data between the GPU and CPU.
Moreover, since this method is only added after pretraining, the LLM can't learn to condition its predictions on the results of arithmetic operations during pretraining.
Since arithmetic is a fundamental building block of more complex tasks, it would be worthwhile to enable LLMs to solve arithmetic tasks directly during pretraining, so that it can use this ability as a subroutine for more complex problems.

\textbf{Our Solution}.
We develop the \textit{Integrated Gated Calculator} (IGC), a module that enables an LLM to accurately perform arithmetic operations, in a single iteration and without using external tools.
It extracts numbers from the tokens in a categorical representation and then emulates the calculation directly on the GPU.

Our contributions are:

\begin{itemize}
    \item \textbf{Innovation}.
    We introduce the \textit{Integrated Gated Calculator}, a novel module that can emulate a calculator (Section~\ref{section-methods}).
    We modify a pretrained Llama 3.1 8B model \citep{touvron2023llama,dubey2024llama} with it and finetune on synthetic data. This enables the LLM to solve complex arithmetic tasks reliably, directly on the GPU, without using COTs, a scratchpad, or calling any tools.
    \item \textbf{Results}.
    We achieve near-perfect generalization on the BigBench Arithmetic Benchmark, outperforming SOTA models that are almost two orders of magnitude larger (Section~\ref{section-experiments}).
    To the best of our knowledge, we are the first to enable an LLM to solve multiplication tasks without the use of external tools or lengthy multi-step procedures.
    \item \textbf{Analysis}.
    We compare our approach with alternatives (Section~\ref{section-comparison-of-methods}) and find that it has numerous advantages besides its strong performance on the benchmark: It is both computationally efficient and interpretable, and it can learn to avoid side effects and destructive interference for problems that do not require arithmetics.
    \item \textbf{Future Work}.
    We describe how the IGC could be integrated into an LLM during pretraining instead of finetuning (Section~\ref{section-future-work}). This would allow the LLM to learn to use it as a subroutine for more complex tasks, an ability that is missing from alternative approaches.
    We further describe how our approach could be generalized and extended to other non-differentiable operations, such as looking up items in a database.
\end{itemize}

\section{Related Work}

\textbf{Word Problems}.
We want to highlight the fact that arithmetic tasks are different from math word problems.
In some cases, models fail to solve word problems even though they follow correct reasoning, because they get the arithmetic operations wrong \citep{schick2023toolformerlanguagemodelsteach,cobbe2021gsm8k,gao2023pal}.
In other cases, models fail to solve word problems with trivial arithmetic operations because they fail to extract the numbers or to format the output correctly. We see examples of this in our analysis of existing benchmark data in Section~\ref{section-experiments}.
In this paper we focus on arithmetic. We discuss in Future Work (Section~\ref{section-future-work}) how our method could be integrated into an LLM more effectively than alternative approaches, which should help greatly with word problems, too.

\textbf{Chain of Thought}.
Chain of Thought methods have shown promising results on a variety of different tasks and for many different models \citep{nye2021show,chung2024scaling}.
\citet{lee2023teachingarithmeticsmalltransformers} investigated ways to teach arithmetic to small transformers and found that COT can help significantly on this task as well.

\textbf{Tool Use}.
\citet{schick2023toolformerlanguagemodelsteach} introduced the Toolformer, a generic method to enable an LLM to interact with an external tool. By interacting with a calculator, this method can perfectly solve arithmetic tasks of any complexity.
Tool use is a very generic technique with applications for many different tasks and domains \citep{qu2024tool}.

\textbf{Other Approaches for Arithmetic Tasks}.
\citet{cobbe2021gsm8k} train verifiers to solve math word problems.
\citet{nye2021show} add scratchpads to the COT approach.
\citet{imani2023mathpromptermathematicalreasoningusing} compares several Chains of Thoughts to improve reliability.
\citet{chen2022program} and \citet{gao2023pal} combine COT and Tool Use by generating executable code.

\textbf{Modifying LLMs}.
Many different techniques for modifying and finetuning LLMs exist \citet{ding2022delta}.
Our approach is most similar to Adapter-based methods \citep{houlsby2019parameter}, which work by injecting a separate smaller neural network into a pretrained LLM. This Adapter module is trained to modify one of the intermediate activations of the LLM, while the base LLM's parameters are kept frozen.
However, our method has several important differences to typical Adapter methods, which we explain in the next Section.

\section{Methods}
\label{section-methods}

\begin{figure*}[t]
    \centering
    \frame{\includegraphics[width=0.279\linewidth]{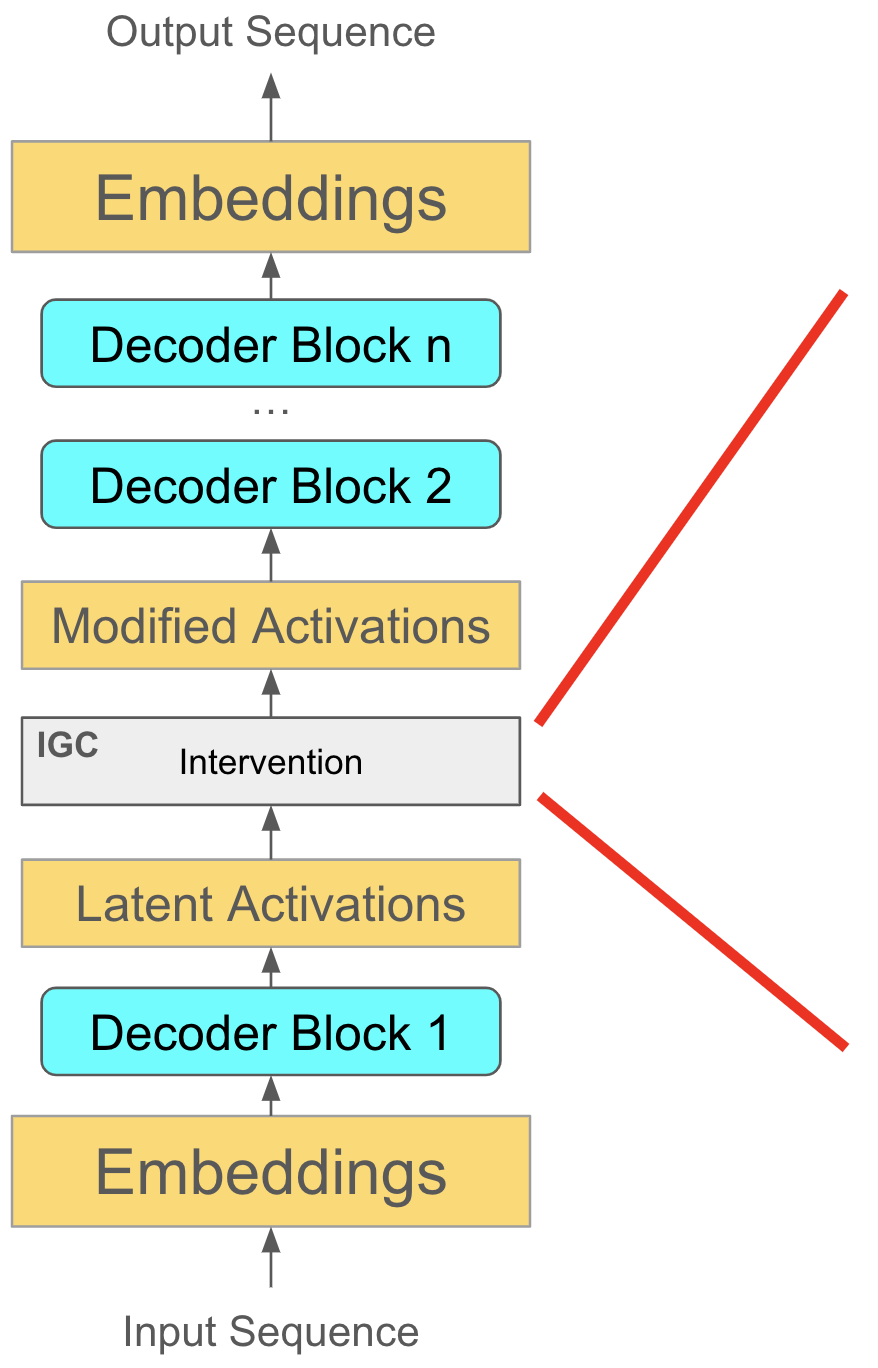}}
    \frame{\includegraphics[width=0.714\linewidth]{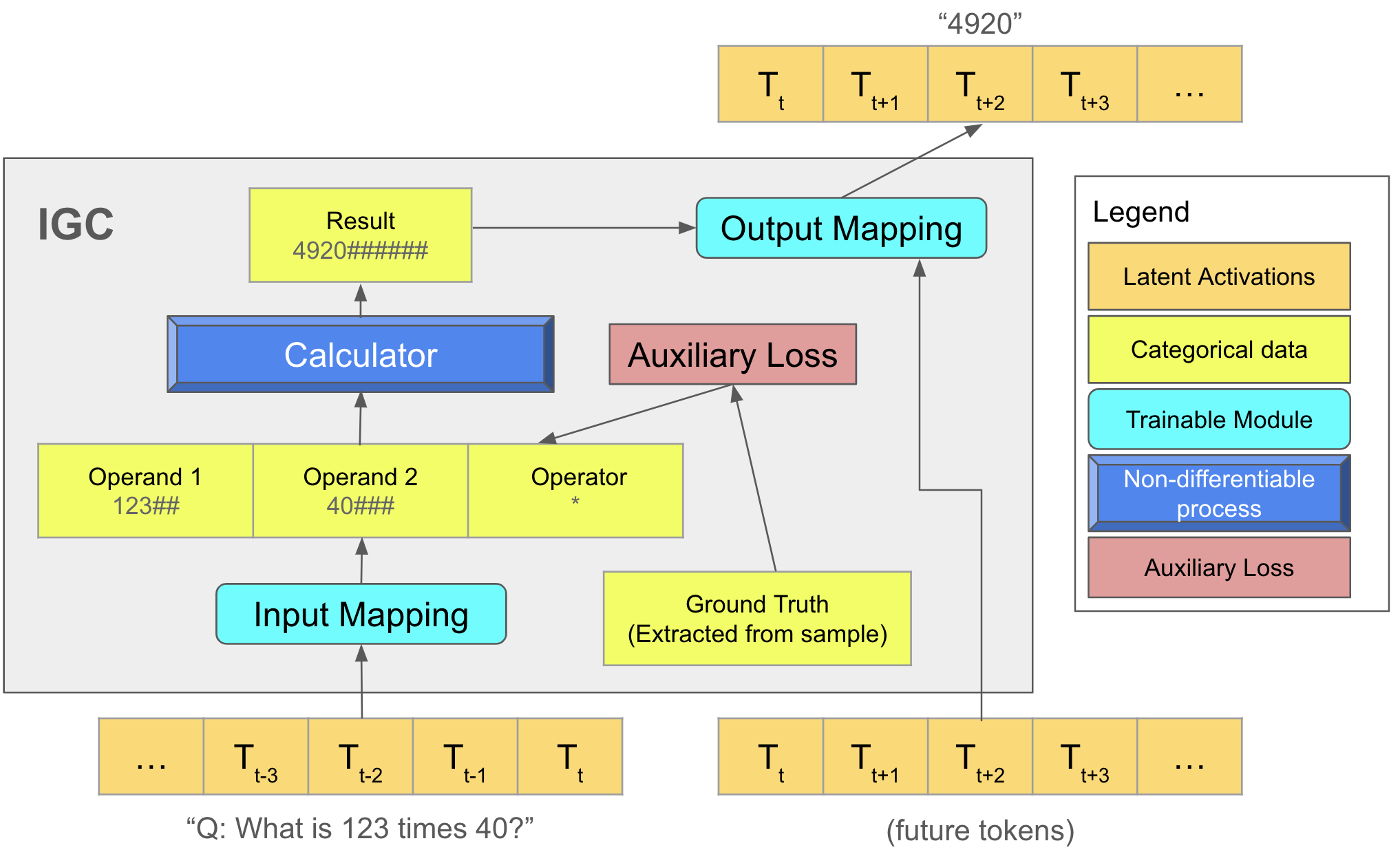}}
    \caption{
    \textbf{Left}. The IGC is inserted into a pretrained LLM after a fixed layer, in this case layer 1. It modifies the output produced by that layer.
    \textbf{Right}. During training, the IGC takes the latent activations produced by the layer as its inputs and splits them into two parts: Before and after the anchor token $T_t$ at time step $t$, which has a special role for argument selection.
    The IGC comprises three components, two of which are trainable submodules:
    The Input Mapping submodule (Figure~\ref{fig:architecture-details}, left) uses the tokens before $T_t$ to extract the arithmetic task from the text and to format it for the calculator. It is trained through an auxiliary loss.
    The calculator itself is emulated on the GPU through a sequence of non-differentiable tensor operations. It is not a trainable component.
    The Output Mapping submodule (Figure~\ref{fig:architecture-details}, right) uses the results of the calculator to modify the tokens after $T_t$. It is trained by the LLM's normal loss function.
    Note that this image shows the training process using teacher forcing. During inference, the Input Mapping and the calculator are executed only on the iteration when the anchor token arrives. Their outputs are cached and reused on subsequent iterations.
    }
    \label{fig:architecture-main}
\end{figure*}

\begin{figure*}[t]
    \centering
    \frame{\includegraphics[width=0.497\linewidth]{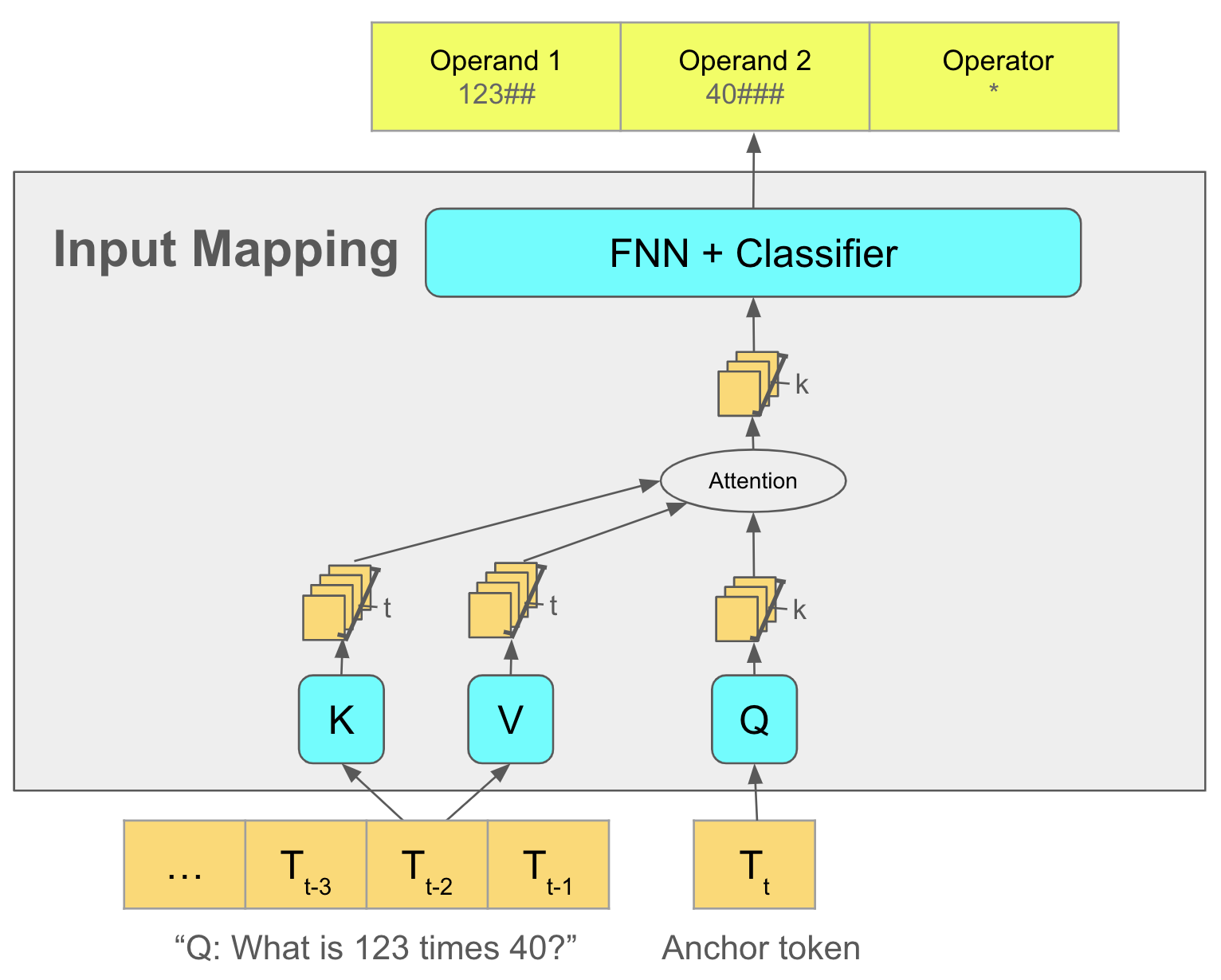}}
    \frame{\includegraphics[width=0.496\linewidth]{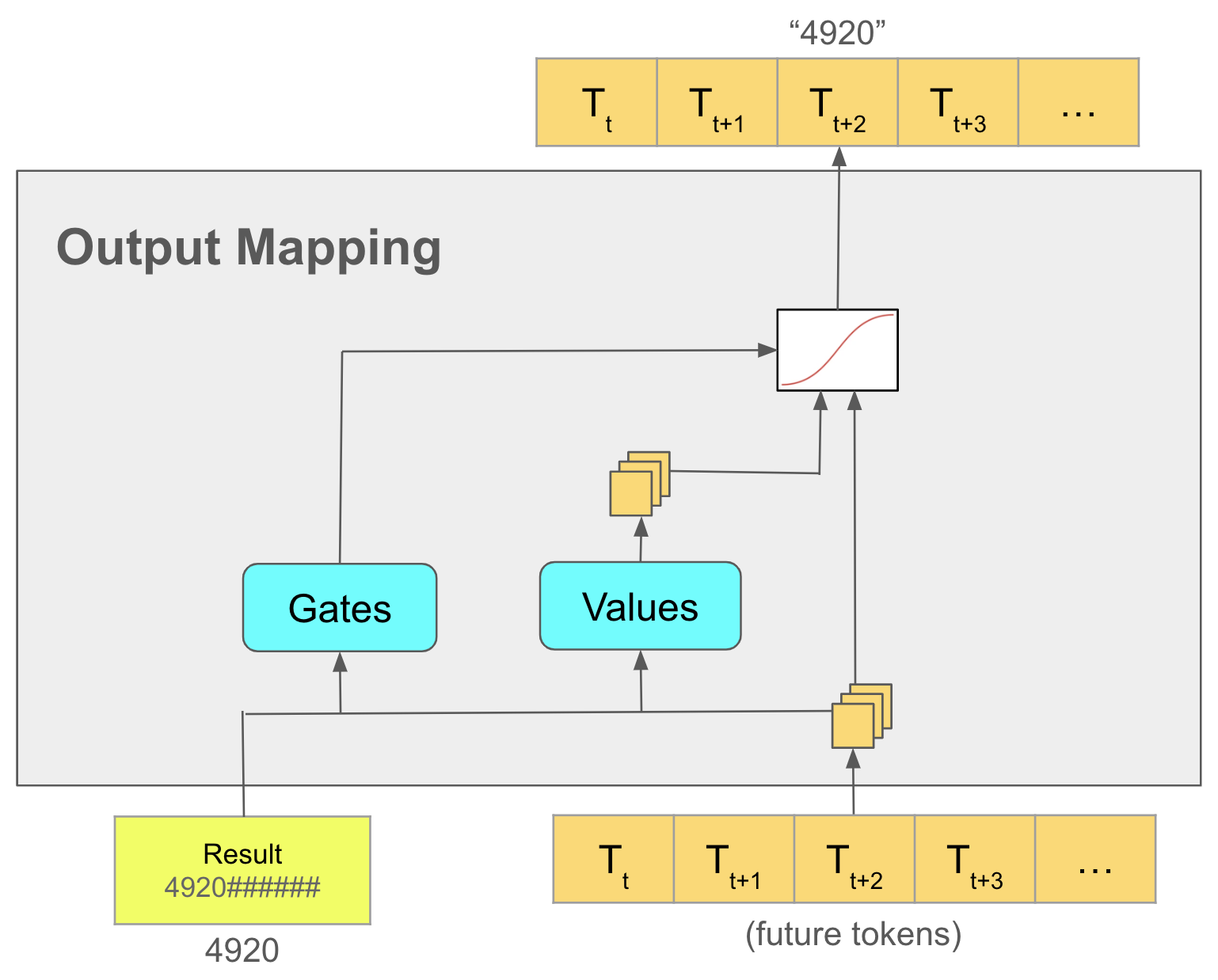}}
    \caption{
    \textbf{Left}. The Input Mapping submodule takes variable-length textual embeddings and extracts the numbers and operator as fixed-length categorical data. The operands and operator are produced as probability distributions over possible digit values for each digit.
    \textbf{Calculator (not shown)}. The calculator discretizes the distributions produced by the Input Mapping submodule by sampling the most probable number and operator. It then emulates the arithmetic operation. The resulting number is formatted using one-hot encoding.
    \textbf{Right}. The Output Mapping submodule uses the fixed-length output of the calculator to modify each of the output tokens. This uses a separate learned gating weight for each token so that it can easily learn to leave tokens unchanged.
    }
    \label{fig:architecture-details}
\end{figure*}

\textbf{Approach}.
Figure~\ref{fig:architecture-main} gives an overview of our approach.
We introduce the Integrated Gated Calculator (IGC), a new module that modifies the output of an existing layer of a pretrained LLM.
We keep the LLM's existing weights frozen and train only the weights of the IGC.
This is similar to Adapter-based tuning methods, but with several important differences:
\begin{itemize}
    \item It has non-differentiable components.
    \item It operates on multiple tokens at once.
    \item It is executed in discrete steps.
    \item It uses gated connections on its outputs.
\end{itemize}
We explain the reasons for and implications of each of these differences in the following.

\subsection{Main Considerations}

\textbf{Non-differentiable Components}.
The IGC uses tensor operations to emulate a calculator directly on the GPU.
The calculator's input and output digits are represented with discrete categorical data, which makes it a non-differentiable operation.
Therefore, the calculator itself is not a trainable component and it blocks the gradient coming from the LLM's main loss.
We therefore have two separate trainable components, which are illustrated in Figure~\ref{fig:architecture-details}: The Output Mapping, which is trained as normal, and the Input Mapping, which does not receive a gradient because of the calculator.
This necessitates a custom training method using an auxiliary loss, which we describe in Section~\ref{section-methods-training}.

\textbf{Dependency on Multiple Tokens}.
Most applications of Adapter-based methods care about abstract concepts like sentiment, or about the presence or absence of specific named entities, since this type of information is often tested in Natural Language Benchmarks such as GLUE \citep{wang2019gluemultitaskbenchmarkanalysis,houlsby2019parameter}.
LLMs can encode such information in a single token or a fixed set of tokens.
However, numbers are encoded in several sequential tokens instead, and their relative position is crucial.
Our architecture needs to reflect this.
We therefore have to learn a mapping from a variable-length set of tokens to the fixed-size input of the calculator.
Conversely, for the output, we need to map the fixed-size number back to all subsequent tokens.
We implement this through attention mechanisms and dynamic gating weights.
We did try a simpler variant for comparison, in which the Input Mapping submodule used fixed inputs, and this architecture performed much worse.

\textbf{Discrete Execution}.
Performing an arithmetic operation is a discrete task and not a distributed heuristic process. One does not perform $5\%$ of a calculation one moment and $\%7$ the next. A calculation is either performed or it is not. Our module is designed to reflect this:
Most adapter methods work by applying a module to the \textit{current} token on \textit{each} iteration.
We instead perform no operations until we are sure that the arithmetic task has been fully described, at time $t$: When the chat switches from the user to the system.
The IGC is then run once, using all tokens available. The token $T_t$ at time $t$ is called the anchor token. The Input Mapping submodule uses it to construct a Query in an attention mechanism, to find all tokens relevant for the arithmetic operation.
During training, we use a form of teacher-forcing described in Section~\ref{section-methods-training}.
During inference, the results of our module at time $t$ are cached for later iterations.
This unusual implementation has several benefits:
\begin{itemize}
    \item \textbf{More Effective Training}.
    We can not be certain what arithmetic operation is needed until all relevant tokens have been encountered. If we try to train the Input Mapping before all of the information it needs is available, it can only learn to guess. This would introduce noise and disrupt the training process.
    \item \textbf{Avoiding Redundancy}.
    If there is only one arithmetic operation, then every iteration after $t$ should learn to perform the exact same Input Mapping. There would be no benefit in repeating the operation multiple times.
\end{itemize}

\textbf{Gated Outputs}.
The Output Mapping submodule uses gated connections with learned and dynamically calculated weights to modify the tokens after $T_t$.
It can learn how much each of the output tokens needs to be modified, and it can simply use near-zero weights to avoid making changes for tasks that do not require arithmetic.
As a consequence, the IGC causes no destructive interference in tasks where it is not needed.
We have experimentally confirmed that this works by testing our architecture on non-arithmetic tasks after training it for the arithmetic tasks described in Section~\ref{section-experiments}.
These tasks used easily recognizable input templates, so confirming it for more complex word problems remains as future work.

\subsection{Training Method}
\label{section-methods-training}

Our training is based on the usual teacher-forcing approach. However, because the calculator is non-differentiable and blocks the gradient to the Input Mapping submodule, we have to slightly modify the algorithm.

\textbf{Ground Truth Data}.
We annotate our training data with ground truth data that specifies for each sample which arithmetic operation needs to be called, and with which operands.
For simple arithmetic templates, this information can easily be extracted automatically.
For more difficult examples, such as word problems, training data can be created by the LLM itself, through a process analogous to the one described in the Toolformer paper \citep{schick2023toolformerlanguagemodelsteach}: The LLM annotates its existing training data with arithmetic operations that it believes would have been helpful. We then measure if annotating the data with the result of that operation reduces perplexity compared to the un-annotated data. If it does, we add the annotation to our training data.

\textbf{Modified Training Process}.
We use this helper information to modify the training process in two ways: Firstly, we apply an auxiliary loss to the Input Mapping component. This is just a simple cross-entropy loss that teaches the Input Mapping submodule to produce the correct input to the calculator. Secondly, we replace the output of the calculator with the correct output, so that the Output Mapping can begin training immediately, before the Input Mapping has converged. This second step is not strictly necessary, but speeds up training.

\textbf{One IGC execution per sample}.
In our training data, each sample requires exactly one use of the calculator, although these can happen at different times for each sample.
This limitation simplifies the training process and allows for more efficient parallelization without loss of generality: Multi-step arithmetic tasks can simply be broken up into individual samples for each step of the calculation.

\subsection{Implementation Details}

\textbf{Effects of Tokenization}.
Previous studies have found that tokenization has a large impact on arithmetic abilities \citep{kim2021have,ding2022delta,liu2023goatfinetunedllamaoutperforms,tweet2024garrethlee}.
For training to be efficient, we need to ensure an inductive bias so that similar tokens can easily be mapped to similar internal representations, for both the input and the output of our module.
The key consideration is that numbers are tokenized from left to right and the calculator's representation of digits must reflect this by being left-aligned: The most significant digit must be assigned to a fixed index, not the least significant one.
All architectures described in this paper use this left-aligned format.
Additionally, we need to consider how digits are chunked into tokens.
While older versions of Llama used one token per digit \citep{yuan2023well,liu2023goatfinetunedllamaoutperforms}, Llama 3.1 \citep{dubey2024llama} groups numbers together in chunks of up to three digits.
This makes it harder for the model to infer the position of each digit, which caused significant problems when we used the more intuitive right-aligned format instead of the left-aligned one.
However, using the left-aligned pattern solved the problem and allowed us to meet and exceed the performance of other approaches.
We expect that the IGC can be adjusted to the tokenization methods and number representations used by other LLMs in a similar manner.
See Section~\ref{section-tokenization} in the Appendix for details.

\textbf{Choosing the Layer}.
We tried applying our module at several different layers of the LLM.
We obtained the best results at early layers.
The results reported in our experiments are all based on applying the module to layer 1.


\section{Experiments}
\label{section-experiments}

\textbf{The BigBench Arithmetic Benchmark}.
In this paper we focus on arithmetic accuracy and not mathematical reasoning in general.
We therefore picked the BigBench Arithmetic benchmark for our evaluations \citep{srivastava2023beyond}.
It uses a deliberately simple template that focuses on raw arithmetic and has a good balance of different operators and input lengths.

\textbf{Alternate Templates}.
To ensure that the simplicity of the BigBench Arithmetic template does not skew results, we additionally trained and tested on several custom templates, as shown in Figure~\ref{fig:training_templates}, in order to make the task more diverse and challenging. We noticed no difference in performance between these templates.

\textbf{Comparisons}.
We modify a pretrained and instruction-finetuned Llama 3.1 8B model with an IGC, train it on synthetic data, and test it on the benchmark.
The best existing results on the benchmark were achieved by PALM 535B \citep{chowdhery2023palm}, which is significantly larger than our model. We therefore also report the performance of relatively smaller models that are no more than one order of magnitude larger than our model.
Unfortunately, a direct comparison with existing benchmark results is slightly unfair, since these were obtained with n-shot prompting instead of finetuning.
We deliberately make things harder for ourselves in order to strengthen the validity of our results: We report the \textit{average} performance of our runs and compare it to the \textit{best} performance of the n-shot runs.
Additionally, we also compare our model to a second baseline, which is based on finetuning: We enhance a Llama model with an Adapter method, using the same parameter count as our model, and finetune it on the same data.
It should also be noted that finetuning more accurately tests for arithmetic abilities than n-shot prompting does. This is evident from inspecting the benchmark results and comparing n-shot results for different values of $n$: There is a lot of variance in performance, even though the arithmetic tasks are exactly the same, and often the 1-shot version outperforms n-shot variants with higher $n$.
The loss of accuracy comes from an inability to understand the formatting required for the output, not an inability to perform the calculation.

\textbf{Results}.
Our model outperforms all baselines by a large margin.
\begin{itemize}
    \item Our model's performance is close to perfect across the board, even for the substantially harder subtask of multiplication.
    \item Our model is much better than the best of the smaller benchmark models.
    \item Compared to PALM 535B, which is optimized for mathematics and almost two orders of magnitude larger, our model is still slightly better for most subtasks and significantly better at multiplication.
    \item We significantly outperform our finetuning-based baseline, which shows that our model's great performance can be attributed to our novel architecture and not to the difference in evaluation methods.
    \item Our experiments had low variance and consistently converged to the reported values for multiple random seeds. In contrast, many of the n-shot benchmarks had high variance and the worst runs of our method still outperformed the best n-shot variants in the benchmark.
\end{itemize}

\textbf{Investigating Anomalies}.
Curiously, our model performs slightly worse at the division subtask than the other operations, even though division is much simpler than multiplication. After investigating the possible causes, we attribute this to the fact that the BigBench Arithmetic benchmark used a different algorithm for generating random operands than we did. The test data therefore follows a different distribution than the training data.

\begin{table*}
    \centering
    \begin{tabular}{|l|c|c|c|c|}
        \hline
        Task & Best small model & PALM 535B & Llama 8B baseline & IGC \\
        \hline
        \#Parameters & < 100B & 535B & 8B+17M & 8B+17M \\
        \hline
        Method & n-shot & n-shot & finetuning & finetuning \\
        \hline
        \hline
        \textbf{Overall} & 0.49 & 0.94 & 0.70 & \textbf{0.99} \\
        \hline
        Addition & 0.49 & 0.94 & 0.95 & \textbf{0.99} \\
        Subtraction & 0.69 & 0.96 & 0.88 & \textbf{0.99} \\
        Multiplication & 0.35 & 0.91 & 0.22 & \textbf{0.99} \\
        Division & 0.71 & 0.97 & 0.75 & \textbf{0.98} \\
        \hline
    \end{tabular}
    \caption{
    The accuracy of different models on the BigBench Arithmetic benchmark. The two columns about n-shot methods were extracted from official benchmark results, 
    while the two finetuned variants were trained by us. We report the averages for the finetuned models and the best value for any $n$ for the n-shot benchmarks. Despite this lopsided comparison, our model still outperforms everything else.
    }
    \label{tab:performance_comparison}
\end{table*}

\begin{figure}
    \centering
    \includegraphics[width=0.95\linewidth]{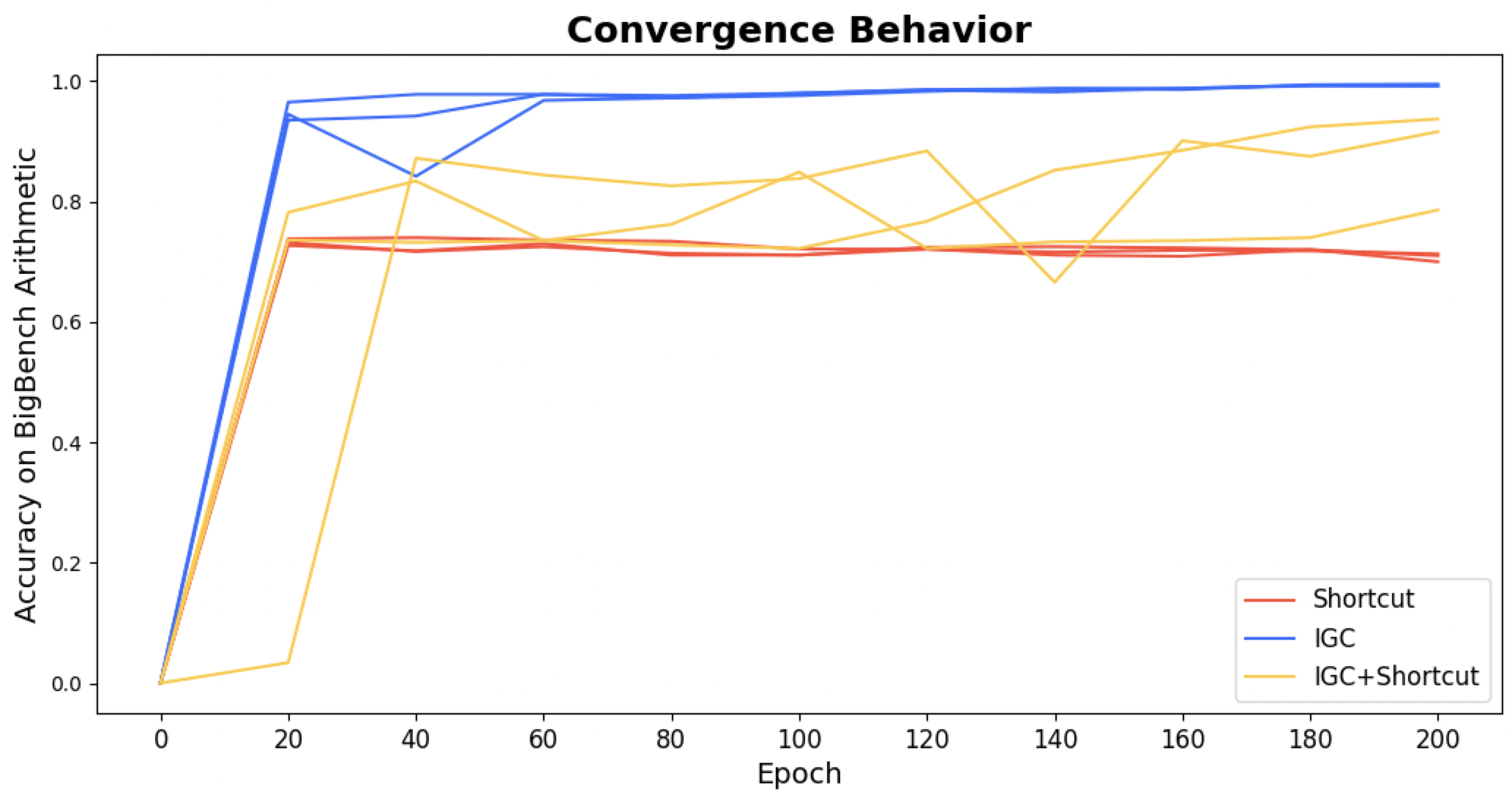}
    \caption{
    The accuracy of various architectures on the BigBench Arithmetic benchmark as training proceeds.
    Multiple lines with the same color correspond to different random seeds for the same architecture.
    }
    \label{fig:convergence-behavior}
\end{figure}

\textbf{Ablations}.
We also created hybrid modules in which the Input Mapping produces an additional output that is given to the Output Mapping submodule directly, making it end-to-end trainable just like a normal finetuning method.
Figure~\ref{fig:convergence-behavior} shows convergence behavior and final performance for several architectures: Our original IGC module without this shortcut, the IGC module with the shortcut, and a baseline that uses only the shortcut and uses no integrated calculator.
All architectures start at zero accuracy because the LLM makes trivial formatting mistakes. They rapidly improve as they learn the template.
The variant that uses only the shortcut showed no improvements after this initial adaptation and its performance is equal to the finetuning baseline. This is unsurprising, since the pretraining likely already included basic arithmetic operations.
The IGC module converges within 70 epochs and its test performance remains stable after convergence.
We observe high variance in the test accuracy for the IGC+shortcut hybrid models, but their final performance is overall lower than the pure IGC without a shortcut connection.
We hypothesize that this is caused by overfitting and destructive interference: The finetuning component converges faster than the calculator, but it does not generalize well.


\textbf{Model Sizes and Efficiency}.
The IGC has 17 million parameters and is integrated into a pretrained Llama model with 8 billion parameters.
Meanwhile, PALM 535B has almost two orders of magnitude more parameters than the Llama model, and four orders of magnitude more than the IGC.
The size of our module is trivial compared to the gain in performance it provides.
The training process is fast and efficient as well: We generated an optimized dataset by filtering out frequently-occurring subsequences of tokens.
Using this technique, our models converged within two to four days of training with a set of only 10,000 samples, on a single GPU.
We note that such a small dataset is not enough for a normal model to learn how to generalize, as evidenced by our ablations.
This shows that our approach is less data hungry than alternative approaches.
We suspect that the reason for this is that the IGC's internal calculator is a perfect emulator, resulting in a much better inductive bias than a randomly initialized neural network.

\section{Comparison to Other Methods}
\label{section-comparison-of-methods}

Table~\ref{tab:comparison_of_traits} shows a high-level comparison between our method and other methods for solving arithmetic tasks.

\textbf{Capability}.
The only method that can solve arbitrary arithmetic tasks as reliably as the IGC is the use of external tools.

\textbf{Efficiency}.
Our method runs in a single step and avoids expensive transfers of data between GPU and CPU, making it more efficient than both COT and tool use.

\textbf{Interpretability}.
The IGC is highly interpretable because the numbers and operator are represented explicitly.
Moreover, the results of the calculation are mapped back to the LLM using learned gates, which allows us to measure how much they affect future tokens.

\textbf{Integration}.
We have reason to believe that our method can be more cleanly integrated into an LLM than other methods.
Firstly, the IGC is entirely \textit{internal} to the model and does not affect output tokens directly. This is important, because the output is an information bottleneck.
If the arithmetic operation is only a subtask of a larger task, needing to generate tokens for it may distract the LLM from its main task.
Additionally, teacher forcing generates separate gradients for each output token, which means that later steps taken for the same arithmetic operations can not repair mistakes made at earlier steps. In contrast, since the IGC is internal to the model and only executed once, it receives a single, coherent gradient.
Secondly, all three approaches (IGC, COT, tool use) share a common weakness, but only the IGC offers a path to fix that weakness:
All three approaches are \textit{added after pretraining}. This implies that the model can not know what the true result of a calculation is during pretraining. It is therefore forced to learn how to make plausible guesses.
Later, when we add the technique to solve arithmetic tasks correctly, the LLM has to unlearn these heuristics again.
The results of our ablation studies in Section~\ref{section-experiments} show that the presence of these incorrect heuristics is harmful and leads to a severe reduction in training speed and generalization ability.
Unlike COT and tool use methods, our method can be modified to avoid these problems:
We describe in Future Work how the IGC could be trained during pretraining (Section~\ref{section-future-work}).

\textbf{Extensibility}.
Our method is specialized for arithmetic and trained on numbers up to a fixed length. By design, it can not help with any other type of task and it needs to be trained up to the largest number length we expect to see.
We address this in the Limitations (Section~\ref{section-limitations}) and explain why it is not much of a hindrance in practice.
We also note that the basic design of our module could be generalized and adjusted to other tasks. We describe how in Future Work (Section~\ref{section-future-work}).

\begin{table*}
    \centering
    \begin{tabular}{|l|l|c|c|c|c|}
        \hline
        \textbf{Feature} & \textbf{Feature Type} & \textbf{Basic LLM} & \textbf{COT} & \textbf{Tool use} & \textbf{IGC} \\
        \hline
        Addition \& Subtraction & Capability & \xmark & \cmark & \cmark & \cmark \\
        Multiplication \& Division & Capability & \xmark & \xmark & \cmark & \cmark \\
        \hline
        Single-step Solutions & Efficiency & \cmark & \xmark & \xmark & \cmark \\
        On GPU & Efficiency & \cmark & \cmark & \xmark & \cmark \\
        \hline
        Explicit Representations & Interpretability & \xmark & \cmark & \cmark & \cmark \\
        Modularity & Interpretability & \xmark & \pmark & \pmark & \cmark \\
        \hline
        Internal & Integration & \cmark & \xmark & \xmark & \cmark \\
        Pretraining & Integration & \cmark & \xmark & \xmark & \cmark \\
        \hline
        Dynamic Number Lengths & Extensibility & \cmark & \cmark & \cmark & \xmark \\
        Generically Extensible & Extensibility & \cmark & \cmark & \cmark & \pmark \\
        \hline
    \end{tabular}
    \caption{
    A comparison of different methods for solving arithmetic tasks.
    \textbf{Addition \& Subtraction}. Can the model generalize on addition and subtraction tasks?
    \textbf{Multiplication \& Division}. Can the model generalize on multiplication and division tasks?
    \textbf{Single-step Solutions}. Does the model run in constant time?
    \textbf{On GPU}. Does the model run entirely on the GPU?
    \textbf{Explicit Representation}. Can you tell when the model performs an arithmetic operation?
    \textbf{Modularity}. Can you tell how the model uses the results of an arithmetic operation?
    \textbf{Internal}. Is the arithmetic task solved entirely within latent variables, or does it need to generate output tokens?
    \textbf{Pretraining}. Is it possible to train the technique during pretraining, or is it only added afterwards?
    \textbf{Dynamic Number Lengths}. Can the model work with inputs of any size?
    \textbf{Generically Extensible}. Can the technique be extended to other tasks?
    }
    \label{tab:comparison_of_traits}
\end{table*}

\section{Future Work}
\label{section-future-work}

\textbf{Using the IGC during Pretraining}.
As explained in Section~\ref{section-comparison-of-methods}, it would be preferable if the IGC was trained directly during pretraining.
The difficulty here is that the IGC requires annotated training data for its auxiliary loss.
Fortunately, it should suffice to intersperse the LLM's normal training data with small amounts of our annotated arithmetic-specific data. During training, the IGC would be executed for all data, but its Input Mapping submodule would only be trained on this annotated data. For data that lacks this annotation, the submodule simply does not receive a gradient.
This makes the training process resilient to noise or even to mistakes in the remainder of the training data.
We note that it would be possible to apply the same technique to pretrain a tool-using model. However, doing so will increase the number of tokens produced for all affected samples, because the LLM needs to produce tokens in order to call tools. This reduces training efficiency and could even be actively harmful to performance because these additional tokens can act as a distractor. In contrast, since the IGC does not produce tokens and uses learned gates on its outputs, it is unlikely to cause any harm even when the dataset is annotated incorrectly.

\textbf{Word Problems}.
Word problems are different from pure arithmetic problems: In addition to the ability to perform arithmetic operations, they also require the model to identify the correct operator and inputs from the text.
The IGC is only designed to perform arithmetic operations, so this is outside of its scope. However, it is relevant to investigate how effectively the IGC can be integrated into the LLM: When an LLM extracts an arithmetic task from a word problem, does it represent this task internally in a consistent manner, so that our Input Mapping submodule can access it effectively?
We expect that this integration will be better if the IGC is trained during pretraining: The Input Mapping submodule generates gradients that encourage the rest of the LLM to extract numbers in an interpretable format, but these gradients are ignored if we only train the IGC and keep the LLM's parameter's frozen.
\textbf{Generalizing the IGC Mechanism}.
The core component of the IGC is a non-differentiable calculator. From the perspective of the trainable components, this is a blackbox.
That raises the question: What other mechanisms could be implemented in such a blackbox?
For example, if we replaced it with a lookup table and adjusted the training mechanism appropriately, it would enable the model to perform database lookups or knowledge graph traversals in a single iteration and without generating any tokens.
Such a mechanism could be used to improve upon the popular technique of Retrieval Augmented Generation \citep{lewis2020retrieval}.

\section{Conclusion}

We introduce the Integrated Gated Calculator (IGC), a module that enhances an LLM with the ability to solve arithmetic tasks.
We achieve near-perfect generalization on the BigBench Arithmetic benchmark, outperforming all existing models.
In addition to its impressive performance, the IGC is also more practical than competing approaches: It avoids both the need for expensive tool calls, as well as lengthy and distracting Chains of Thought.
We discuss how to integrate our module into an LLM during pretraining and explain why this could improve the model's ability even further, supported by empirical evidence from our ablation studies.
Lastly, we note that our method could be generalized to integrate other types of non-differentiable tools into an LLM.

\section{Limitations}
\label{section-limitations}

\textbf{Fixed Maximum Length}.
By construction, the IGC uses a fixed maximum input length for its numbers. This size can be arbitrarily large, but can not be adjusted later.
To mitigate this issue, we can simply train the IGC with the largest size that we expect to see for the majority of practical tasks.
That number may be surprisingly small: In the MATH dataset \citep{hendrycksmath2021}, a standard benchmark for math word problems, 99\% of numbers have four digits or less.
In the rare cases when the model does have to deal with larger numbers, the LLM can still use the IGC as a very reliable approximator.
It should also be noted that the IGC is not mutually exclusive with other methods: If the model encounters an arithmetic task that is both too large for the IGC \textit{and} that requires high accuracy, it can resort to calling external tools.
\textit{This mirrors human behavior}: We learn to solve numbers up to a certain size in our heads. If we encounter tasks more complex than that, we either perform a rough calculation, or we resort to tools.
Being able to solve simple arithmetic tasks in our heads without needing to use a tool is useful, but starting at a certain level of complexity it is no longer worth the effort.

\section{Acknowledgements}

This work was supported by a grant by the NHR Verein (National High Performance Computing, Germany).

\bibliography{main}

\begin{thebibliography}{29}
\providecommand{\natexlab}[1]{#1}

\bibitem[{Achiam et~al.(2023)Achiam, Adler, Agarwal, Ahmad, Akkaya, Aleman, Almeida, Altenschmidt, Altman, Anadkat et~al.}]{achiam2023gpt}
Josh Achiam, Steven Adler, Sandhini Agarwal, Lama Ahmad, Ilge Akkaya, Florencia~Leoni Aleman, Diogo Almeida, Janko Altenschmidt, Sam Altman, Shyamal Anadkat, et~al. 2023.
\newblock Gpt-4 technical report.
\newblock \emph{arXiv preprint arXiv:2303.08774}.

\bibitem[{bench authors(2023)}]{srivastava2023beyond}
BIG bench authors. 2023.
\newblock \href {https://openreview.net/forum?id=uyTL5Bvosj} {Beyond the imitation game: Quantifying and extrapolating the capabilities of language models}.
\newblock \emph{Transactions on Machine Learning Research}.

\bibitem[{Brown(2020)}]{brown2020language}
Tom~B Brown. 2020.
\newblock Language models are few-shot learners.
\newblock \emph{arXiv preprint arXiv:2005.14165}.

\bibitem[{Chen et~al.(2022)Chen, Ma, Wang, and Cohen}]{chen2022program}
Wenhu Chen, Xueguang Ma, Xinyi Wang, and William~W Cohen. 2022.
\newblock Program of thoughts prompting: Disentangling computation from reasoning for numerical reasoning tasks.
\newblock \emph{arXiv preprint arXiv:2211.12588}.

\bibitem[{Chowdhery et~al.(2023)Chowdhery, Narang, Devlin, Bosma, Mishra, Roberts, Barham, Chung, Sutton, Gehrmann et~al.}]{chowdhery2023palm}
Aakanksha Chowdhery, Sharan Narang, Jacob Devlin, Maarten Bosma, Gaurav Mishra, Adam Roberts, Paul Barham, Hyung~Won Chung, Charles Sutton, Sebastian Gehrmann, et~al. 2023.
\newblock Palm: Scaling language modeling with pathways.
\newblock \emph{Journal of Machine Learning Research}, 24(240):1--113.

\bibitem[{Chung et~al.(2024)Chung, Hou, Longpre, Zoph, Tay, Fedus, Li, Wang, Dehghani, Brahma et~al.}]{chung2024scaling}
Hyung~Won Chung, Le~Hou, Shayne Longpre, Barret Zoph, Yi~Tay, William Fedus, Yunxuan Li, Xuezhi Wang, Mostafa Dehghani, Siddhartha Brahma, et~al. 2024.
\newblock Scaling instruction-finetuned language models.
\newblock \emph{Journal of Machine Learning Research}, 25(70):1--53.

\bibitem[{Cobbe et~al.(2021)Cobbe, Kosaraju, Bavarian, Chen, Jun, Kaiser, Plappert, Tworek, Hilton, Nakano, Hesse, and Schulman}]{cobbe2021gsm8k}
Karl Cobbe, Vineet Kosaraju, Mohammad Bavarian, Mark Chen, Heewoo Jun, Lukasz Kaiser, Matthias Plappert, Jerry Tworek, Jacob Hilton, Reiichiro Nakano, Christopher Hesse, and John Schulman. 2021.
\newblock Training verifiers to solve math word problems.
\newblock \emph{arXiv preprint arXiv:2110.14168}.

\bibitem[{Ding et~al.(2022)Ding, Qin, Yang, Wei, Yang, Su, Hu, Chen, Chan, Chen et~al.}]{ding2022delta}
Ning Ding, Yujia Qin, Guang Yang, Fuchao Wei, Zonghan Yang, Yusheng Su, Shengding Hu, Yulin Chen, Chi-Min Chan, Weize Chen, et~al. 2022.
\newblock Delta tuning: A comprehensive study of parameter efficient methods for pre-trained language models.
\newblock \emph{arXiv preprint arXiv:2203.06904}.

\bibitem[{Dubey et~al.(2024)Dubey, Jauhri, Pandey, Kadian, Al-Dahle, Letman, Mathur, Schelten, Yang, Fan et~al.}]{dubey2024llama}
Abhimanyu Dubey, Abhinav Jauhri, Abhinav Pandey, Abhishek Kadian, Ahmad Al-Dahle, Aiesha Letman, Akhil Mathur, Alan Schelten, Amy Yang, Angela Fan, et~al. 2024.
\newblock The llama 3 herd of models.
\newblock \emph{arXiv preprint arXiv:2407.21783}.

\bibitem[{Dziri et~al.(2024)Dziri, Lu, Sclar, Li, Jiang, Lin, Welleck, West, Bhagavatula, Le~Bras et~al.}]{dziri2024faith}
Nouha Dziri, Ximing Lu, Melanie Sclar, Xiang~Lorraine Li, Liwei Jiang, Bill~Yuchen Lin, Sean Welleck, Peter West, Chandra Bhagavatula, Ronan Le~Bras, et~al. 2024.
\newblock Faith and fate: Limits of transformers on compositionality.
\newblock \emph{Advances in Neural Information Processing Systems}, 36.

\bibitem[{Gao et~al.(2023)Gao, Madaan, Zhou, Alon, Liu, Yang, Callan, and Neubig}]{gao2023pal}
Luyu Gao, Aman Madaan, Shuyan Zhou, Uri Alon, Pengfei Liu, Yiming Yang, Jamie Callan, and Graham Neubig. 2023.
\newblock Pal: Program-aided language models.
\newblock In \emph{International Conference on Machine Learning}, pages 10764--10799. PMLR.

\bibitem[{Garreth(2024)}]{tweet2024garrethlee}
Lee Garreth. 2024.
\newblock Tweet.
\newblock https://x.com/garrethleee/status/1860039446311371132.
\newblock A tweet about a paper, not yet published as of the time of this writing.

\bibitem[{Hendrycks et~al.(2021)Hendrycks, Burns, Kadavath, Arora, Basart, Tang, Song, and Steinhardt}]{hendrycksmath2021}
Dan Hendrycks, Collin Burns, Saurav Kadavath, Akul Arora, Steven Basart, Eric Tang, Dawn Song, and Jacob Steinhardt. 2021.
\newblock Measuring mathematical problem solving with the math dataset.
\newblock \emph{NeurIPS}.

\bibitem[{Houlsby et~al.(2019)Houlsby, Giurgiu, Jastrzebski, Morrone, De~Laroussilhe, Gesmundo, Attariyan, and Gelly}]{houlsby2019parameter}
Neil Houlsby, Andrei Giurgiu, Stanislaw Jastrzebski, Bruna Morrone, Quentin De~Laroussilhe, Andrea Gesmundo, Mona Attariyan, and Sylvain Gelly. 2019.
\newblock Parameter-efficient transfer learning for nlp.
\newblock In \emph{International conference on machine learning}, pages 2790--2799. PMLR.

\bibitem[{Imani et~al.(2023)Imani, Du, and Shrivastava}]{imani2023mathpromptermathematicalreasoningusing}
Shima Imani, Liang Du, and Harsh Shrivastava. 2023.
\newblock \href {https://arxiv.org/abs/2303.05398} {Mathprompter: Mathematical reasoning using large language models}.
\newblock \emph{Preprint}, arXiv:2303.05398.

\bibitem[{Kim et~al.(2021)Kim, Hong, Kim, Kang, and Myaeng}]{kim2021have}
Jeonghwan Kim, Giwon Hong, Kyung-min Kim, Junmo Kang, and Sung-Hyon Myaeng. 2021.
\newblock Have you seen that number? investigating extrapolation in question answering models.
\newblock In \emph{Proceedings of the 2021 Conference on Empirical Methods in Natural Language Processing}, pages 7031--7037.

\bibitem[{Lee et~al.(2023)Lee, Sreenivasan, Lee, Lee, and Papailiopoulos}]{lee2023teachingarithmeticsmalltransformers}
Nayoung Lee, Kartik Sreenivasan, Jason~D. Lee, Kangwook Lee, and Dimitris Papailiopoulos. 2023.
\newblock \href {https://arxiv.org/abs/2307.03381} {Teaching arithmetic to small transformers}.
\newblock \emph{Preprint}, arXiv:2307.03381.

\bibitem[{Lewis et~al.(2020)Lewis, Perez, Piktus, Petroni, Karpukhin, Goyal, K{\"u}ttler, Lewis, Yih, Rockt{\"a}schel et~al.}]{lewis2020retrieval}
Patrick Lewis, Ethan Perez, Aleksandra Piktus, Fabio Petroni, Vladimir Karpukhin, Naman Goyal, Heinrich K{\"u}ttler, Mike Lewis, Wen-tau Yih, Tim Rockt{\"a}schel, et~al. 2020.
\newblock Retrieval-augmented generation for knowledge-intensive nlp tasks.
\newblock \emph{Advances in Neural Information Processing Systems}, 33:9459--9474.

\bibitem[{Liu and Low(2023)}]{liu2023goatfinetunedllamaoutperforms}
Tiedong Liu and Bryan Kian~Hsiang Low. 2023.
\newblock \href {https://arxiv.org/abs/2305.14201} {Goat: Fine-tuned llama outperforms gpt-4 on arithmetic tasks}.
\newblock \emph{Preprint}, arXiv:2305.14201.

\bibitem[{McLeish et~al.(2024)McLeish, Bansal, Stein, Jain, Kirchenbauer, Bartoldson, Kailkhura, Bhatele, Geiping, Schwarzschild et~al.}]{mcleish2024transformers}
Sean McLeish, Arpit Bansal, Alex Stein, Neel Jain, John Kirchenbauer, Brian~R Bartoldson, Bhavya Kailkhura, Abhinav Bhatele, Jonas Geiping, Avi Schwarzschild, et~al. 2024.
\newblock Transformers can do arithmetic with the right embeddings.
\newblock \emph{arXiv preprint arXiv:2405.17399}.

\bibitem[{Nye et~al.(2021)Nye, Andreassen, Gur-Ari, Michalewski, Austin, Bieber, Dohan, Lewkowycz, Bosma, Luan et~al.}]{nye2021show}
Maxwell Nye, Anders~Johan Andreassen, Guy Gur-Ari, Henryk Michalewski, Jacob Austin, David Bieber, David Dohan, Aitor Lewkowycz, Maarten Bosma, David Luan, et~al. 2021.
\newblock Show your work: Scratchpads for intermediate computation with language models.
\newblock \emph{arXiv preprint arXiv:2112.00114}.

\bibitem[{Qu et~al.(2024)Qu, Dai, Wei, Cai, Wang, Yin, Xu, and Wen}]{qu2024tool}
Changle Qu, Sunhao Dai, Xiaochi Wei, Hengyi Cai, Shuaiqiang Wang, Dawei Yin, Jun Xu, and Ji-Rong Wen. 2024.
\newblock Tool learning with large language models: A survey.
\newblock \emph{arXiv preprint arXiv:2405.17935}.

\bibitem[{Schick et~al.(2023)Schick, Dwivedi-Yu, Dessì, Raileanu, Lomeli, Zettlemoyer, Cancedda, and Scialom}]{schick2023toolformerlanguagemodelsteach}
Timo Schick, Jane Dwivedi-Yu, Roberto Dessì, Roberta Raileanu, Maria Lomeli, Luke Zettlemoyer, Nicola Cancedda, and Thomas Scialom. 2023.
\newblock \href {https://arxiv.org/abs/2302.04761} {Toolformer: Language models can teach themselves to use tools}.
\newblock \emph{Preprint}, arXiv:2302.04761.

\bibitem[{Thawani et~al.(2021)Thawani, Pujara, Szekely, and Ilievski}]{thawani2021representing}
Avijit Thawani, Jay Pujara, Pedro~A Szekely, and Filip Ilievski. 2021.
\newblock Representing numbers in nlp: a survey and a vision.
\newblock \emph{arXiv preprint arXiv:2103.13136}.

\bibitem[{Thoppilan et~al.(2022)Thoppilan, De~Freitas, Hall, Shazeer, Kulshreshtha, Cheng, Jin, Bos, Baker, Du et~al.}]{thoppilan2022lamda}
Romal Thoppilan, Daniel De~Freitas, Jamie Hall, Noam Shazeer, Apoorv Kulshreshtha, Heng-Tze Cheng, Alicia Jin, Taylor Bos, Leslie Baker, Yu~Du, et~al. 2022.
\newblock Lamda: Language models for dialog applications.
\newblock \emph{arXiv preprint arXiv:2201.08239}.

\bibitem[{Touvron et~al.(2023)Touvron, Lavril, Izacard, Martinet, Lachaux, Lacroix, Rozi{\`e}re, Goyal, Hambro, Azhar et~al.}]{touvron2023llama}
Hugo Touvron, Thibaut Lavril, Gautier Izacard, Xavier Martinet, Marie-Anne Lachaux, Timoth{\'e}e Lacroix, Baptiste Rozi{\`e}re, Naman Goyal, Eric Hambro, Faisal Azhar, et~al. 2023.
\newblock Llama: Open and efficient foundation language models.
\newblock \emph{arXiv preprint arXiv:2302.13971}.

\bibitem[{Wang et~al.(2019)Wang, Singh, Michael, Hill, Levy, and Bowman}]{wang2019gluemultitaskbenchmarkanalysis}
Alex Wang, Amanpreet Singh, Julian Michael, Felix Hill, Omer Levy, and Samuel~R. Bowman. 2019.
\newblock \href {https://arxiv.org/abs/1804.07461} {Glue: A multi-task benchmark and analysis platform for natural language understanding}.
\newblock \emph{Preprint}, arXiv:1804.07461.

\bibitem[{Wei et~al.(2023)Wei, Wang, Schuurmans, Bosma, Ichter, Xia, Chi, Le, and Zhou}]{wei2023chainofthoughtpromptingelicitsreasoning}
Jason Wei, Xuezhi Wang, Dale Schuurmans, Maarten Bosma, Brian Ichter, Fei Xia, Ed~Chi, Quoc Le, and Denny Zhou. 2023.
\newblock \href {https://arxiv.org/abs/2201.11903} {Chain-of-thought prompting elicits reasoning in large language models}.
\newblock \emph{Preprint}, arXiv:2201.11903.

\bibitem[{Yuan et~al.(2023)Yuan, Yuan, Tan, Wang, and Huang}]{yuan2023well}
Zheng Yuan, Hongyi Yuan, Chuanqi Tan, Wei Wang, and Songfang Huang. 2023.
\newblock How well do large language models perform in arithmetic tasks?
\newblock \emph{arXiv preprint arXiv:2304.02015}.

\end{thebibliography}

\appendix

\section{Tokenization}
\label{section-tokenization}

\textbf{The Problem}.
We experimented with different ways to construct the input and output format of our module to improve learning speed and generalization.
The basic goal is that similar tokens must be mapped to similar internal representations, for both the input and the output of our module.
The main hindrance to this is that tokenization groups multiple digits together into single tokens.

\textbf{Analysis}.
We have empirically analyzed the way tokenization works in our model:
Llama 3.1 8B uses a tokenization method that consistently parses numbers from left to right and groups digits together in groups of three, unless there are only two or one digit left. It also conveniently tokenizes in such a way that these 3-digit tokens do not contain non-numeric characters.
When read from the left, the tokens are always multiples of 3 digits, which makes it predictable which part of which token corresponds to which significant digit. For example, the 4th most significant digit always maps to the second token. Only the three least significant digits have any ambiguity.
In contrast, when you read from the right, then figuring out which token a digit belongs to depends on the length of the number, which is non-local information and therefore much harder to learn.
Table~\ref{tab:tokenization_examples} illustrates this.


\textbf{Consequences for the Architecture}.
The left-aligned format requires more work to implement because the position of the digit no longer matches the digit's significance.
Adjusting the architecture to work with left-aligned numbers requires two small changes:
Firstly, each of the digits now needs to be represented as a classifier with eleven options instead of ten: One for each possible digits and one for the special placeholder symbol (an asterisk in Table~\ref{tab:tokenization_examples}).
Secondly, extracting the number from a left-aligned representation requires an additional step compared to the right-aligned representation: We need to know the index of the first placeholder and shift the number accordingly to ensure we get the right magnitude.
This is demonstrated by the code in Section~\ref{appendix-code}.

\textbf{Results}.
We tried a right-aligned version of our architecture as well. The difference in performance was significant and in many cases the right-aligned variant failed to converge or to improve upon the performance of a finetuned model at all.

\textbf{Other Tokenization Methods}.
The left-aligned architecture described here works well because of the way Llama 3.1 tokenizes the data.
Other tokenization models may have different requirements and work better with other types of architectures.
We have written code to automatically analyze the distribution of tokens when generating training data. This code can help you to determine the optimal architecture to use. We will make it available upon acceptance of the paper.

\begin{table*}
    \centering
    \begin{tabular}{|l|l|r|l|}
        \hline
        Number & Tokenization & Right-aligned & Left-aligned \\
        \hline
        1234567890 & "123"456"789"0" & \texttt{\underline{123}456\underline{789}0} & \texttt{\underline{123}456\underline{789}0} \\
        123456789 & "123"456"789" & \texttt{0\underline{123}456\underline{789}} & \texttt{\underline{123}456\underline{789}*} \\
        12345678 & "123"456"78" & \texttt{00\underline{123}456\underline{78}} & \texttt{\underline{123}456\underline{78*}*} \\
        234567890 & "234"567"890" & \texttt{0\underline{234}567\underline{890}} & \texttt{\underline{234}567\underline{890}*} \\
        34567890 & "345"678"90" & \texttt{00\underline{345}678\underline{90}} & \texttt{\underline{345}678\underline{90*}*} \\
        \hline
    \end{tabular}
    \caption{
    Examples of numbers, how they get tokenized, and two different variants for representing them internally. Both variants use a fixed size of 10 digits for illustration. The right-aligned variant is the intuitive, default format, and puts the \emph{least} significant digit at a fixed index. In contrast, the left-aligned format puts the \emph{most} significant digit at a fixed index. The right-aligned format can be padded with zeros, but the left-aligned format requires a special symbol to mark where the number ends.
    The underlining in the two formatted numbers indicates which digits get grouped together into the same token. Note that the underlining is the same for all examples in the left-aligned format, but is inconsistent in the default format. This makes the mapping between the tokens and the format much easier to learn for the left-aligned format.
    The last two lines illustrates that removing digits from the left instead of the right as in the examples above does not lead to the reverse phenomenon: The tokenization is now different, so it doesn't help that the right-aligned representation is similar.
    }
    \label{tab:tokenization_examples}
\end{table*}

\section{Code}
\label{appendix-code}


In the following we provide python code for the three stages of our module.

\subsection{Code of the Input Mapping and Auxiliary Loss}
\label{appendix-code-input-and-auxiliary-loss}

(We will provide code upon acceptance of the paper. The code in its current stage still needs to be optimized and documented for other researchers.)

\subsection{Code for Emulating a Calculator}
\label{appendix-code-emulator}

The calculator emulation works in three steps:
\begin{itemize}
    \item Translate left-aligned digits to numbers. We take inputs in a fixed-length categorical format that represents digits and translate this into a single number.
    \item Perform a calculation on the numbers. To ensure that the module can be trained on batches of data, we perform all four operations in parallel and then take a weighted average.
    \item Translate the result back into left-aligned digits.
\end{itemize}

\textbf{Numerical Accuracy}.
One important implementation detail to be aware of is the numerical accuracy of torch tensors. We only calculate up to 10 digits because that is all that is needed for the benchmark. If the calculator should be able to handle longer numbers that do not fit into a torch.int64 datatype, the calculation must be broken down into several smaller steps. This leads to code that is longer and harder to understand, although the operations are still fairly simple.

(We will provide code upon acceptance of the paper. The code in its current stage still needs to be optimized and documented for other researchers.)

\subsection{Code of the Output Mapping}
\label{appendix-code-output}


(We will provide code upon acceptance of the paper. The code in its current stage still needs to be optimized and documented for other researchers.)


\end{document}